\documentclass[10pt,twocolumn,letterpaper]{article}

\usepackage{cvpr}              % To produce the CAMERA-READY version

\usepackage{cases}
\usepackage{graphicx}
\usepackage{booktabs}
\usepackage{multirow}
\usepackage{makecell}
\usepackage{array}
\usepackage{float}

\definecolor{cvprblue}{rgb}{0.21,0.49,0.74}
\usepackage[pagebackref,breaklinks,colorlinks,allcolors=cvprblue]{hyperref}

\def\paperID{*****} % *** Enter the Paper ID here
\def\confName{CVPR}
\def\confYear{2026}

\title{General Incomplete Multimodal Learning via Dynamic Quality Perception}

\author{
    Xiangyu Meng, 
    Shicai Wei\thanks{Corresponding author.}
    \\
    University of Electronic Science and Technology of China
    \\
    \texttt{fivemeng3@gmail.com}, 
    \texttt{shicaiwei@uestc.edu.cn}
}

\begin{document}
\maketitle
\begin{abstract}

Multimodal learning robust to missing modalities is essential for real-world applications. Existing methods mainly focus on inter-modality missing, where entire modalities are absent, while overlooking intra-modality degradation, where modalities are present but severely corrupted. In practice, these two types of missing often coexist, making existing approaches ineffective. To address this limitation, we propose General Incomplete Multimodal Learning (GIML), a unified framework that simultaneously handles both inter-modality missing and intra-modality degradation through dynamic quality perception. Specifically, GIML models heterogeneous missing patterns as continuous modality information degradation, enabling degradation-aware adaptive fusion.  To achieve reliable quality perception, we introduce a Noise-aware Quality Estimator that learns the mapping from corrupted features to noise intensity through controlled noise injection. Furthermore, we propose a Noise–Semantic Decoupled module that separates semantic information from noise interference. This improves robustness and generalization to unseen corruption patterns. Extensive experiments across datasets with diverse modality types demonstrate the effectiveness and generality of GIML. Code is available at: \url{https://github.com/Yu-Five/GIML}. 

% \keywords{Multimodal learning \and Dynamic quality perception  \and Modality missing}

\end{abstract}    
\section{Introduction}
\label{sec:intro}

Multimodal learning has achieved strong performance across many vision tasks~\cite{liang2021multibench, sun2024generative, wei2024gradient}. However, most existing methods assume that all modalities are available during training and inference. This assumption often fails in practice due to device limitations~\cite{liu2021face, pinto2015face} or challenging operating conditions~\cite{li2021dynamic, stroud2020d3d}. Such incompleteness can significantly harm model performance and robustness, motivating the study of incomplete multimodal learning.

Existing approaches generally fall into two categories: imputation-based and joint representation-based methods.
Imputation-based methods reconstruct missing modalities from available ones at the input~\cite{cai2018deep, jue2019integrating, liu2021face} or representation level~\cite{dai2025unbiased, zhang2025incomplete, araujo2025cav, jin2025rohydr, zhu2025proxy, sun2024redcore, wang2023learnable, sikdar2025ogp}. While this type of method is straightforward, their performance heavily depends on reconstruction quality and errors can propagate to downstream tasks. Thus, joint representation-based methods instead learn unified embeddings from different modality combinations without explicit reconstruction. They capture the shared feature for all possible input modality combinations to address the incomplete multimodal learning issue~\cite{havaei2016hemis, zhang2022mmformer, wei2024robust, dai2024study, wei2023mmanet}. 

Despite the success of existing methods, most focus on inter-modality missing, where entire modalities are absent. In practice, however, modality incompleteness also occurs within modalities, where inputs are present but corrupted by noise. Recent works such as T2DR~\cite{lin2025t2dr} and TMDC~\cite{zhuang2025tmdc} attempt to address this issue, but they treat intra-modality corruption and inter-modality missing as two separate problems and handle them sequentially. This could encounter the optimization conflicts across stages, resulting in sub-optimal performance. Moreover, they are typically designed for specific intra-modal corruption, such as Gaussian degradation. This limits their robustness to unseen missing patterns.

To address these limitations, we propose General Incomplete Multimodal Learning (GIML), a unified framework that simultaneously handles intra-modality corruption and inter-modality missing through dynamic quality perception. Specifically, GIML models heterogeneous missing patterns as continuous modality degradation and estimates modality quality to guide adaptive fusion. As degradation increases from mild corruption to complete absence, the modality weight smoothly decays from 1 to 0, naturally recovering the conventional missing-modality setting.  Besides, we propose a Noise–Semantic Decoupled module that separates semantic information from noise interference. This design encourages learning noise-invariant semantic representations, improving robustness to diverse and previously unseen corruption patterns while facilitating more reliable degradation estimation.

Accurate quality estimation under severe noise remains challenging. Existing approaches, such as energy score \cite{liu2020energy} and probabilistic embedding \cite{shi2019probabilistic}, often produce unreliable estimates, assigning non-negligible weights to heavily corrupted modalities. To address this issue, we introduce a Noise-aware Quality Estimator (NQE) that learns a direct mapping from corrupted features to noise intensity via controlled noise injection, enabling reliable quality estimation across the full degradation spectrum.

Overall, our contributions can be summarized as follows:
\begin{itemize}
    \item We propose a general incomplete multimodal learning framework (GIML) that unifies intra-modality and inter-modality missing through dynamic quality perception, enabling their joint optimization within a single stage.

     \item  We propose a Noise-Semantic Decoupled module that explicitly disentangles semantic-related information from noise-related components, guiding task optimization to rely on noise-invariant semantic representations. This improves the robustness of GIML to unseen noise corruption.

    \item We propose a noise-aware quality estimator that adds controlled noise to the modality data and predicts its intensity from the modality feature, allowing accurate uncertainty measures under various noise conditions.

    \item Extensive experiments on multiple datasets demonstrate the effectiveness of the proposed GIML to handle the intra-modality and inter-modality missing simultaneously.
\end{itemize}

\section{Related Work}
\label{sec:related_work}

\subsection{Incomplete multimodal learning}

Multimodal learning leverages diverse information to improve performance in many tasks, but real-world data is often incomplete, which can degrade models. Existing approaches fall into two categories.

\textbf{Imputation-based methods} aim to recover missing modalities form available ones. Early works \cite{cai2018deep, jue2019integrating, liu2021face} reconstruct missing data in the input space, which is challenging due to data complexity. To mitigate this issue, later works reconstruct missing modality features in the latent space \cite{dai2025unbiased, zhang2025incomplete, araujo2025cav, jin2025rohydr,wei2023mshnet}, or estimate missing modality features by exploiting cross-modal correlations among available modalities \cite{zhu2025proxy, sun2024redcore, wang2023learnable, sikdar2025ogp}. Although effective, imputation-based methods remain sensitive to reconstruction accuracy, and errors may propagate to downstream tasks. 

\textbf{Joint representation-based methods} aim to learn robust representations from available modalities without explicit reconstruction. Many enforce modality-invariant embeddings through shared subspaces and modality dropout \cite{havaei2016hemis, zhang2022mmformer, wei2024robust, dai2024study}, which improves robustness but may reduce modality-specific cues. Others exploit complementary information via cross-modal attention \cite{zhang2024towards, praveen2024recursive, zhuanghyper} or dynamic weighting \cite{li2025simmlm, xu2024leveraging}. However, multimodal representation learning itself inherently suffers from modality imbalance \cite{wang2020makes, peng2022balanced, chaudhuri2025closer,wei2025boosting,wei2025improving} due to factors such as disparities in information richness, optimization dynamics. When missing rates across modalities are unequal, this intrinsic imbalance is further amplified as the disparities between modalities widen.

Recent efforts have been made to handle both intra-modality and inter-modality incompleteness. T2DR \cite{lin2025t2dr} and TMDC \cite{zhuang2025tmdc} adopt two-stage pipelines: first mitigating intra-modality degradation, then addressing inter-modality missing. These approaches, however, fail to handle simultaneous intra-modality and inter-modality missing and rely on representations entangled with noise, which limits generalization to unseen corruption.

\subsection{Dynamic Multimodal Fusion}

Multimodal dynamic fusion adaptively weights modalities within a joint representation to integrate multi-source information. Existing work mainly differs in how fusion weights are estimated.

\textbf{Saliency-Based Methods.} These methods compute fusion weights from unimodal salience or cross-modal interactions, implemented via linear projections \cite{wu2025beyond,wei2026unbiased}, attention mechanisms \cite{li2024crossfuse, lv2025rethinking}, or structured routing such as Mixture-of-Experts (MoE) \cite{huai2025cl, jing2025evomoe, li2025simmlm, han2024fusemoe, cao2023multi}. Because salience typically correlates with information strength, adaptive weighting tends to favor dominant modalities, causing the fused representation to underutilize complementary but less expressive cues.

\textbf{Uncertainty-Based Methods.} These approaches regulate modality contributions through uncertainty estimation, broadly categorized as sampling-based and representation-based methods. Sampling-based approaches, such as Monte Carlo Dropout \cite{zeevi2025rate, djupskaas2025unreliable}, approximate epistemic uncertainty via multiple stochastic forward passes but incur additional computation. Representation-based approaches model modality features as probability distributions, where dispersion reflects uncertainty. Variational methods \cite{yang2024uncertainty, harrison2024variational} learn distribution parameters and sample features through reparameterization, while probabilistic embeddings \cite{venkataramanan2025probabilistic, lin2025intra, shi2019probabilistic} directly parameterize modality embeddings as distributions. Without ground-truth uncertainty supervision, these methods rely on indirect proxies such as output variance or auxiliary losses, which may not faithfully reflect prediction reliability.

\section{Method}

\begin{figure*}[tb] 
  \centering
  \includegraphics[width=1.0\textwidth]{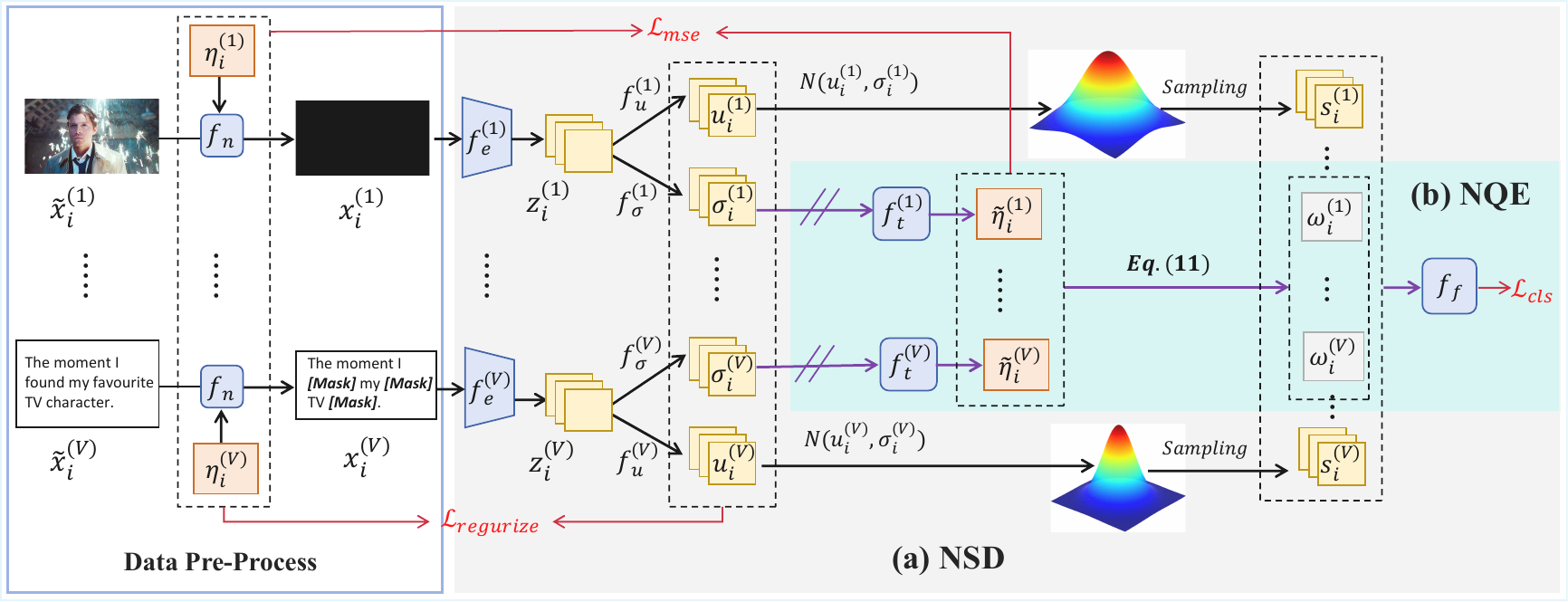} 
  \caption{Overall framework of the proposed GIML. It consists of two components: a Noise-Semantic Decoupled (NSD) module and a Noise-aware Quality Estimator (NQE). NSD disentangles semantic information from noise-related components to enhance robustness under diverse and unseen corruption patterns, then NQE estimates modality quality to guide adaptive fusion. Additionally, "//" denotes gradient detach operations. }
  % \vspace{-3.5mm}
  \label{fig:framework}
\end{figure*}

In this work, we propose GIML, a general quality-aware framework for incomplete multimodal learning. It unifies intra-modality and inter-modality missing as continuous modality quality degradation. As illustrated in Fig. \ref{fig:framework}, GIML consists of two components: a Noise-Semantic Decoupled (NSD) module and a Noise-aware Quality Estimator (NQE). NSD disentangles semantic information from noise-related components at the representation level, preventing noisy features from contaminating task-relevant semantics. This decoupling encourages the model to rely on noise-invariant representations, thereby improving robustness under varying and previously unseen corruption patterns. Meanwhile, NQE formulates modality uncertainty as data degradation and calibrates it via controlled noise injection to learn reliable quality estimates. The estimated quality then modulates fusion weights, ensuring that modality contributions decay smoothly with increasing degradation.

\noindent \textbf{Notations}. The main notations used in this work are defined as follows. Let $\mathcal{D} = \{(X_i, y_i)\}_{i=1}^N$ be a multimodal dataset containing $N$ samples, where each sample $X_i = (\tilde x_i^{(1)}, \dots, \tilde x_i^{(V)})$ consists of $V$ modality inputs and $y_i \in \{1, \dots, M\}$ is its label. The $v$-th modality encoder is represented by $f_e^{(v)}$. The noise injection function $f_n$ applies degradation with intensity $\eta$ such as Gaussian noise, mask noise, or saltpepper noise to clean modality inputs. The degradation estimator $f_t^{(v)}$ converts uncertainty into a comparable score for fusion weighting, and $f_f$ computes the final fused multimodal representation.

% $\eta$

\subsection{General Incomplete Multimodal Learning}

We first introduce the conventional incomplete multimodal learning,  which primarily focuses on inter-modality missing, where certain modalities are entirely absent. Then, we extend it to a more general scenario, where modalities could be partially missing.

Specifically, the multimodal embedding $\tilde z_i^{\tau}$ of $X_i$ in conventional incomplete multimodal learning can be expressed as follows,

\begin{numcases}{}
\label{drop}
   \tilde z_i^{\tau}=f_{f}(\theta_{f},\tilde {z_i}^{(1)}*{\delta_i}^{(1)},...,\tilde {z_i}^{(V)}*{\delta_i}^{(V)})\\
  \tilde {z_i}^{(v)}=f_{e}^{(v)}({\theta_{e}}^{v},\tilde {x_i}^{v}), \quad v\in[1,V]
\end{numcases} where ${r_i}^{(v)}$ is the embedding of the $v_{th}$ modality. $\theta_{f}$ is the parameter for the fusion module. ${\theta_{e}}^{v}$ is the parameters for the $v_{th}$ modality encoder. $ {\delta_i}^{v}  \in \{0,1\}$ is the  Bernoulli indicator for the $v_{th}$ modality of $x_i$. It is randomly set to either 0 or 1 to simulate random modality missing. This makes the model robust for inter-modality missing during inference.

However, in real-world applications, modality inputs are more likely to suffer varying degrees of quality degradation than to be absent, such as impulse-noise contamination or Gaussian blurring. As a result, modality incompleteness often appears in a continuous manner, which cannot be adequately characterized by a binary indicator. To simulate such scenarios, we introduce a noise injection function $f_n$ that applies degradation with intensity $\eta_i^{(v)}$ to clean modality inputs:
\begin{equation}
{x}_i^{(v)} = f_n(\tilde x_i^{(v)}, \eta_i^{(v)}),
\end{equation}

Here, $\eta_i^{(v)}$ is a relative calibration signal that controls degradation severity, rather than a physical noise label. $f_n$ produces various types of corruption based on $\eta_i^{(v)}$, with larger values indicating heavier degradation, thus spanning from clean to complete absence.

To this end, we extend the conventional binary indicator ${\delta_i}^v \in \{0,1\}$ to a continuous coefficient $w_i^{(v)} \in [0,1]$ that reflects modality quality. Accordingly, the multimodal embedding of $x_i$ in general incomplete multimodal learning is formulated as,

\begin{numcases}{}
\label{drop}
   z_i^{\tau}=f_{f}(\theta_{f},{z_i}^{(1)}*{w_i}^{(1)},...,{z_i}^{(V)}*{w_i}^{(V)})\\
  {z_i}^{(v)}=f_{e}^{(v)}({\theta_{e}}^{v},{x}_i^{(v)}), \quad v\in[1,V]
\end{numcases} 

 Here, $w_i^{(v)}=1$ indicates a fully reliable modality, while $w_i^{(v)}=0$ corresponds to a completely missing modality. Values $w_i^{(v)} \in (0,1)$ represent partially degraded modalities, where smaller values indicate more severe corruption. In this way, modality degradation can be modeled as a continuous transition from slight corruption to complete absence, enabling a unified formulation of intra-modality degradation and inter-modality missing.

Although GIML enables a unified treatment of modality degradation, it also changes the inference paradigm. Instead of simply checking modality availability, GIML must estimate modality reliability and assign adaptive fusion weights. This introduces two key challenges. First, the model must learn noise-robust representations that generalize across diverse and unseen corruption patterns. Second, it must accurately quantify modality uncertainty and translate it into the weighting coefficient $w$ for each modality. To address these challenges, we first introduce the Noise-Semantic Decoupled (NSD) module to enhance robustness to diverse noise interference, and then propose the Noise-aware Quality Estimator (NQE) to accurately estimate modality quality for adaptive fusion.

\subsection{Noise-Semantic Decoupling}
\label{sec:NSD}

As discussed above, GIML needs to generalize across diverse and unseen corruption patterns. However, most existing incomplete multimodal learning methods rely on deterministic embeddings, which implicitly entangle task-relevant semantics with degradation patterns. As a result, the learned representation tends to capture corruption-specific characteristics, leading to overfitting and poor generalization to unseen noise conditions. To address this issue, we propose a Noise–Semantic Decoupled (NSD) representation that models modality features as probabilistic distributions rather than deterministic vectors. This design enables an explicit separation between task-relevant semantic information and degradation-induced uncertainty, preventing semantic representations from being contaminated by noise patterns.

For simplicity, we define the probabilistic embedding obeys a multivariate Gaussian distribution. Specifically, given the modality embedding $z_i^{(v)}$, we parameterize a Gaussian distribution:
\begin{equation}
\mu_i^{(v)} = f_\mu^{(v)}(z_i^{(v)}), \quad
\log \sigma_i^{(v)} = f_\sigma^{(v)}(z_i^{(v)}).
\end{equation}

where the mean $\mu_i^{(v)}$ encodes task-relevant semantic information, while the variance $\sigma_i^{(v)}$ represents degradation-induced uncertainty.

Now, the representation of each sample becomes a stochastic embedding sampled from $\mathcal{N}(\mu_i^{(v)}, (\sigma_i^{(v)})^2 \mathbf{I})$. Nevertheless, the sampling operation is not differentiable. Thus, we consider the reparameterization trick~\cite{wei2024robust} to enable backpropagation:
\begin{equation}
s_i^{(v)} = \mu_i^{(v)} + \sigma_i^{(v)} \odot \epsilon, 
\quad \epsilon \sim \mathcal{N}(0,\mathbf{I}).
\end{equation}

Moreover, to explicitly enforce the separation between semantics and degradation, we introduce two complementary constraints. First, a cross-entropy classification loss $\mathcal{L}_{cls}^{(v)}$ is applied to the sampled representation $s_i^{(v)}$, encouraging the mean $\mu_i^{(v)}$ to preserve discriminative semantic information even under stochastic perturbations. Second, we introduce a degradation-aware prior:
\begin{equation}
\mathcal{L}_{reg}^{(v)} =
\frac{1}{N} \sum_{i=1}^{N}
KL\Big[
\mathcal{N}(\mu_i^{(v)}, (\sigma_i^{(v)})^2 \mathbf{I})
\;\|\;
\mathcal{N}(0, (\eta_i^{(v)})^2 \mathbf{I})
\Big].
\end{equation}

This prior aligns the predicted variance $\sigma_i^{(v)}$ with the degradation intensity $\eta_i^{(v)}$, preventing degradation-related variations from being absorbed into the semantic mean. The prior mean is set to zero to avoid imposing additional semantic bias on $\mu$. Consequently, semantic information and degradation uncertainty are captured in different statistical dimensions, allowing the model to learn noise-invariant semantic representations and improving robustness to diverse and unseen noise conditions.

\subsection{Noise-aware Quality Estimator}
\label{sec:NQE}

The proposed NSD module improves robustness to diverse noise patterns by decoupling semantic information from noise interference. However, effective fusion also requires accurately estimating the severity of modality degradation to assign appropriate weights. However, existing approaches typically measure modality reliability using unsupervised statistics derived from model predictions~\cite{liu2020energy} or feature distributions~\cite{shi2019probabilistic}, which often become misaligned with the true degradation level when noise is either too weak or too severe. To address this limitation, we propose a lightweight Noise-aware Quality Estimator (NQE) $f_t^{(v)}$, which learns to predict the intrinsic noise intensity of each modality through controlled perturbations, enabling accurate and reliable quality estimation across the full degradation spectrum.

Specifically, NQE establishes an explicit mapping from the uncertainty 
$\sigma_i^{(v)}$ produced by NSD to the degradation intensity:

\begin{equation}
\hat{\eta}_i^{(v)} = f_t^{(v)}(\sigma_i^{(v)}).
\end{equation}

Unlike prior methods that rely on implicit reliability inference, 
NQE is trained with direct supervision using the ground-truth degradation intensity $\eta_i^{(v)}$:
\begin{equation}
\mathcal{L}_{mse}^{(v)} =
\frac{1}{N} \sum_{i=1}^{N}
\left(
\hat{\eta}_i^{(v)} - \eta_i^{(v)}
\right)^2.
\end{equation}

The calibrated degradation estimates $\hat{\eta}_i^{(v)}$ are then used 
to guide quality-aware fusion. Following the inverse-variance principle established in ARL \cite{wei2025improving}, we compute adaptive modality weights as

\begin{equation}
\omega_i^{(v)} =
\frac{
V \cdot \sum_{u \neq v} (\hat{\eta}_i^{(u)})^2
}{
\sum_{k=1}^{V} (\hat{\eta}_i^{(k)})^2
},
\end{equation}

Since the degradation scores are explicitly calibrated to the true corruption levels, modalities with higher estimated degradation are adaptively down-weighted, 
while more reliable modalities contribute more to the fused representation. 
This explicit calibration facilitates stable fallback to missing-modality modeling.

Finally, the overall training objective of GIML integrates all loss terms:
\begin{equation}
\mathcal{L}
=
\mathcal{L}_{cls}^{f}
+
\beta_1 \sum_{k=1}^V \mathcal{L}_{cls}^{(k)}
+
\beta_2 \mathcal{L}_{reg}
+
\beta_3 \mathcal{L}_{mse}.
\end{equation}
Here, $\mathcal{L}_{cls}^{f}$ denotes the multimodal task loss computed on the fused representation, and $\mathcal{L}_{cls}^{(k)}$ represents the unimodal classification loss for the $k$-th modality. $\mathcal{L}_{reg}$ is the regularization term introduced in the NSD module, while $\mathcal{L}_{mse}$ supervises the noise intensity prediction in the noise-aware quality estimator.

\section{Experiment}

\subsection{Datasets}
\noindent \textbf{CREMA-D} \cite{cao2014crema} is an audio-visual dataset for emotion recognition, containing 7,442 video clips labeled with six basic emotions. The dataset is split into 6,698 training samples and 744 testing samples.

\noindent \textbf{Kinetics-Sounds (KS)} \cite{arandjelovic2017look} is a subset of the Kinetics dataset focusing on human actions observable both visually and aurally. It consists of 19,000 ten-second video clips spanning 34 action categories, with 15,000 clips for training, 1,900 for validation, and 1,900 for testing.

\noindent \textbf{MVSA-Single} \cite{niu2016sentiment} is a multimodal sentiment analysis dataset of social media posts, containing paired images and text. Each sample is labeled with a positive, negative, or neutral sentiment. In this work, we follow the train/validation/test split defined in \cite{zhang2023provable} to evaluate cross-modal sentiment recognition.

\noindent \textbf{MOSI} \cite{zadeh2018multi} is a benchmark dataset for multimodal sentiment analysis, containing audio, visual, and textual modalities. It includes 2,199 video clips extracted from 93 monologue videos, each annotated with a sentiment score ranging from -3 (strongly negative) to 3 (strongly positive). For evaluation, we adopt a binary sentiment setting, considering only positive versus negative labels, to test the model’s ability to handle multimodal information in real-world scenarios.

\noindent \textbf{NVGesture} \cite{gupta2016online} is a large-scale hand gesture dataset with 25 gesture classes performed by 20 participants under varying lighting and background conditions. The dataset contains multiple modalities captured by a Kinect sensor. In our experiments, we follow the official train/validation/test split and use the RGB and Depth modalities for multimodal learning.

\begin{table}[t]
\caption{\footnotesize Performance comparison on CREMA-D and KS under varying intra-modality degradation ratios and inter-modality missing settings. `Intra ratio $(r_a, r_v)$' indicates Mask-based intra-modality missing in each modality, and `inter' specifies which modalities are available during testing. The best results are highlighted in bold.}
% \vspace{-1mm}
\label{tab:experiments_av_mask}
\centering
\scriptsize
\setlength{\tabcolsep}{0.1pt}
\renewcommand{\arraystretch}{1.2}

\begin{tabular}{
>{\centering\arraybackslash}p{1.0cm}
|>{\centering\arraybackslash}p{1.5cm}
|>{\centering\arraybackslash}p{0.6cm}
|>{\centering\arraybackslash}p{0.9cm}
>{\centering\arraybackslash}p{0.9cm}
|>{\centering\arraybackslash}p{0.9cm}
>{\centering\arraybackslash}p{0.9cm}
|>{\centering\arraybackslash}p{0.9cm}
>{\centering\arraybackslash}p{0.9cm}
}

\hline
\multirow{2}{*}{Dataset}
& \multirow{2}{*}{\makecell{intra ratio\\($r_a$, $r_v$)}}
& \multirow{2}{*}{inter}
& \multicolumn{2}{c|}{TMDC}
& \multicolumn{2}{c|}{T2DR}
& \multicolumn{2}{c}{GIML} \\
\cline{4-9}
& & & Acc & F1 & Acc & F1 & Acc & F1 \\
\hline

% ================= CREMAD-D =================
\multirow{10}{*}{\rotatebox{90}{CREMAD-D}}

& (0.0,0.0) & \multirow{6}{*}{AV}
& 66.99 & 67.03 & 67.93 & 67.82 & \textbf{73.94} & \textbf{74.08} \\
& (0.5,0.5) &
& 61.69 & 61.64 & 62.69 & 62.67 & \textbf{67.80} & \textbf{67.78} \\
& (0.1,0.5) &
& 61.85 & 61.85 & 63.82 & 63.90 & \textbf{68.18} & \textbf{68.13} \\
& (0.5,0.1) &
& 66.37 & 66.28 & 67.45 & 67.37 & \textbf{72.64} & \textbf{72.64} \\
& (0.3,0.7) &
& 57.23 & 57.35 & 59.89 & 60.04 & \textbf{62.93} & \textbf{62.80} \\
& (0.7,0.3) &
& 65.35 & 65.31 & 65.16 & 65.19 & \textbf{67.34} & \textbf{67.37} \\
\cline{2-9}

& (0.0,-) & \multirow{2}{*}{A}
& 52.66 & 52.69 & 53.92 & 54.15 & \textbf{55.57} & \textbf{55.13} \\
& (0.5,-) &
& 48.60 & 48.40 & \textbf{50.38} & \textbf{50.34} & 50.24 & 50.07 \\
\cline{2-9}

& (--,0.0) & \multirow{2}{*}{V}
& 46.80 & 44.54 & 48.95 & 48.96 & \textbf{60.38} & \textbf{60.21} \\
& (--,0.5) &
& 41.68 & 41.28 & 42.63 & 40.95 & \textbf{55.30} & \textbf{54.72} \\
\hline
\hline
% ================= KS =================
\multirow{10}{*}{\rotatebox{90}{KS}}

& (0.0,0.0) & \multirow{6}{*}{AV}
& 60.40 & 60.54 & 62.73 & 62.39 & \textbf{69.51} & \textbf{69.22} \\
& (0.5,0.5) &
& 58.08 & 57.90 & 60.31 & 60.11 & \textbf{63.54} & \textbf{63.06} \\
& (0.1,0.5) &
& 57.67 & 57.49 & 59.62 & 59.38 & \textbf{65.20} & \textbf{64.74} \\
& (0.5,0.1) &
& 60.55 & 60.44 & 62.70 & 62.32 & \textbf{66.76} & \textbf{66.40} \\
& (0.3,0.7) &
& 56.75 & 56.68 & 59.03 & 58.79 & \textbf{61.68} & \textbf{61.26} \\
& (0.7,0.3) &
& 58.97 & 58.78 & 60.72 & 60.24 & \textbf{63.47} & \textbf{63.15} \\
\cline{2-9}

& (0.0,-) & \multirow{2}{*}{A}
& 38.67 & 38.25 & 41.21 & 41.21 & \textbf{45.13} & \textbf{44.66} \\
& (0.5,-) &
& 37.24 & 36.74 & 36.88 & 36.95 & \textbf{41.27} & \textbf{40.82} \\
\cline{2-9}

& (--,0.0) & \multirow{2}{*}{V}
& 46.43 & 44.63 & 44.90 & 43.25 & \textbf{55.80} & \textbf{55.22} \\
& (--,0.5) &
& 43.04 & 41.77 & 40.85 & 39.80 & \textbf{50.67} & \textbf{50.17} \\
\hline
\end{tabular}
% \vspace{-3mm}
\end{table}

\subsection{Experimental settings}

\noindent \textbf{Dataset preprocessing.} For CREMA-D, a single frame is randomly sampled per video, and audio is loaded at 22,050 Hz. KS uses three frames per video, with audio at 16,000 Hz. MOSI samples five frames uniformly per video, and NVGesture samples 24 frames uniformly from both RGB and Depth streams. All visual frames are resized to 224×224. For visual and audio modalities, we adopt ResNet \cite{he2016deep} as the backbone, and for text we use BERT \cite{devlin2019bert} with the first six layers frozen.  Training data contains 50\% clean samples and 50\% samples with random mask rate ranging from 0.0 to 1.0 at intervals of 0.1, simulating diverse real-world inputs.

\noindent \textbf{Missingness setting.} In our experiments, we introduce two types of missingness: Mask and Gaussian. Mask sets a proportion of pixels to zero. In extreme cases, this results in a fully black image, simulating complete visual or audio dropout. All performance comparison experiments are conducted with mask noise. Besides, we also introduce  Gaussian noise as the unseen noise to study the generalization ability of GIML.

\noindent \textbf{Training.} All models are trained using Stochastic Gradient Descent (SGD) with a momentum of 0.9 and a weight decay of 1e-4, with an initial learning rate of 1e-3. Batch sizes are set to 64 for CREMA-D and KS, and 32 for the other datasets. During training, model selection is performed based on the average validation accuracy on both clean samples and those with median noise intensity. Accuracy and weighted F1-score are used as the primary evaluation metrics, denoted as Acc (\%) and F1 (\%) in the tables, respectively.

\subsection{Comparison with State-of-the-art Methods}
% CRAMED-D和KS
% 前面

% MVSA_Single
\begin{table}[t]
\caption{\footnotesize MVSA-Single performance under different intra-modality and inter-modality missing settings.  The best results are highlighted in bold.}
% \vspace{-2mm}
\label{tab:experiments_vt_mask}
\centering
\scriptsize % 字号 10pt
\setlength{\tabcolsep}{2pt} % 列间距
\renewcommand{\arraystretch}{1.2} % 行距

\begin{tabular}{
>{\centering\arraybackslash}p{1.2cm}  % IAM radio
|>{\centering\arraybackslash}p{0.6cm}  % Inter
|>{\centering\arraybackslash}p{0.9cm}  % TMDC Acc
>{\centering\arraybackslash}p{0.9cm}  % TMDC F1
|>{\centering\arraybackslash}p{0.9cm}  % T2DR Acc
>{\centering\arraybackslash}p{0.9cm}  % T2DR F1
|>{\centering\arraybackslash}p{0.9cm}  % GIML Acc
>{\centering\arraybackslash}p{0.9cm} % GIML F1
}

\hline
\multirow{2}{*}{\makecell{intra ratio\\($r_t$, $r_v$)}} 
& \multirow{2}{*}{inter} 
& \multicolumn{2}{c}{TMDC} 
& \multicolumn{2}{|c}{T2DR} 
& \multicolumn{2}{|c}{GIML} \\
\cline{3-8}
& & Acc & F1 & Acc & F1 & Acc & F1 \\
\hline

% tv 合并 6 行
(0.0, 0.0) & \multirow{6}{*}{VT} & 77.54 & 76.51 & 76.56 & 75.65 & \textbf{77.73} & \textbf{77.36} \\
(0.5, 0.5) & & 68.28 & 67.51 & 69.57 & 69.17 & \textbf{71.37} & \textbf{70.29} \\
(0.1, 0.5) & & 76.17 & 75.46 & 75.39 & 74.67 & \textbf{76.68} & \textbf{76.00} \\
(0.5, 0.1) & & 68.52 & 67.24 & 69.30 & 67.94 & \textbf{70.98} & \textbf{70.04} \\
(0.3, 0.7) & & 73.09 & 72.37 & 72.93 & \textbf{72.42} & \textbf{73.28} & 72.28 \\
(0.7, 0.3) & & 63.12 & 61.57 & 63.09 & 61.96 & \textbf{66.02} & \textbf{64.71} \\
\hline

% t 合并 2 行
(0.0, -) & \multirow{2}{*}{T} & 77.73 & 76.63 & 75.20 & 74.04 & \textbf{78.48} & \textbf{78.21} \\
(0.5, -) & & 69.53 & 68.55 & 69.22 & 68.20 & \textbf{70.47} & \textbf{69.58} \\
\hline

% v 合并 2 行
(-, 0.0) & \multirow{2}{*}{V} & 49.41 & 38.84 & 50.78 & \textbf{45.81} & \textbf{51.17} & 42.60 \\
(-, 0.5) & & 48.87 & 40.25 & \textbf{50.94} & 37.66 & 50.78 & \textbf{48.50} \\
\hline
\end{tabular}
\end{table}

% NVGesture
\begin{table}[t]
\caption{\footnotesize Performance on NVGesture under different intra-modality and inter-modality settings.  Best results are highlighted in bold. }
% \vspace{-2mm}
\label{tab:experiments_rd_mask}
\centering
\scriptsize % 字号 10pt
\setlength{\tabcolsep}{2pt} % 列间距
\renewcommand{\arraystretch}{1.2} % 行距

\begin{tabular}{
>{\centering\arraybackslash}p{1.6cm}  % IAM radio
|>{\centering\arraybackslash}p{0.6cm}  % Inter
|>{\centering\arraybackslash}p{0.9cm}  % TMDC Acc
>{\centering\arraybackslash}p{0.9cm}  % TMDC F1
|>{\centering\arraybackslash}p{0.9cm}  % T2DR Acc
>{\centering\arraybackslash}p{0.9cm}  % T2DR F1
|>{\centering\arraybackslash}p{0.9cm}  % GIML Acc
>{\centering\arraybackslash}p{0.9cm} % GIML F1
}

\hline
\multirow{2}{*}{\makecell{intra ratio\\($r_{rgb}$, $r_{depth}$)}} 
& \multirow{2}{*}{inter} 
& \multicolumn{2}{c}{TMDC} 
& \multicolumn{2}{|c}{T2DR} 
& \multicolumn{2}{|c}{GIML} \\
\cline{3-8}
& & Acc & F1 & Acc & F1 & Acc & F1 \\
\hline

% tv 合并 9 行
(0.0, 0.0) & \multirow{6}{*}{RD} & 48.13 & 49.03 & 46.43 & 47.48 & \textbf{56.22} & \textbf{55.79} \\
(0.5, 0.5) & & 47.84 & 48.56 & 46.60 & 47.71 & \textbf{52.99} & \textbf{54.26} \\
(0.1, 0.5) & & 47.76 & 48.73 & 46.76 & 47.53 & \textbf{56.51} & \textbf{57.37} \\
(0.5, 0.1) & & 48.34 & 49.33 & 47.68 & 48.69 & \textbf{55.39} & \textbf{55.98} \\
(0.3, 0.7) & & 48.30 & 49.41 & 46.10 & 47.13 & \textbf{53.94} & \textbf{55.47} \\
(0.7, 0.3) & & 48.09 & 49.14 & 47.10 & 47.98 & \textbf{53.78} & \textbf{54.53} \\
\hline

% t 合并 3 行
(0.0, -) & \multirow{2}{*}{R} & 40.21 & 40.43 & 38.22 & 38.68 & \textbf{52.49} & \textbf{52.52} \\
(0.5, -) & & 39.63 & 39.88 & 38.67 & 38.97 & \textbf{44.27} & \textbf{45.35} \\
\hline

% v 合并 3 行
(-, 0.0) & \multirow{2}{*}{D} & 38.42 & 35.64 & 36.97 & 34.28 & \textbf{51.16} & \textbf{51.78} \\
(-, 0.5) & & 38.42 & 35.55 & 38.76 & 35.73 & \textbf{51.12} & \textbf{52.01} \\
\hline
\end{tabular}
% \vspace{-2mm}
\end{table}

% MOSI
\begin{table}[t]
\caption{\footnotesize CMU-MOSI performance under various intra- and inter-modality degradation settings. Best results are highlighted in bold.}
% \vspace{-2mm}
\label{tab:experiments_avt_mask}
\centering
\scriptsize % 字号调整为 10pt，和规范表格一致
\setlength{\tabcolsep}{2pt} % 调整列间距
\renewcommand{\arraystretch}{1.2} % 行距略微增加

\begin{tabular}{
>{\centering\arraybackslash}p{1.8cm} 
|>{\centering\arraybackslash}p{0.6cm} 
|>{\centering\arraybackslash}p{0.9cm}
>{\centering\arraybackslash}p{0.9cm}
|>{\centering\arraybackslash}p{0.9cm}
>{\centering\arraybackslash}p{0.9cm}
|>{\centering\arraybackslash}p{0.9cm}
>{\centering\arraybackslash}p{0.9cm}
}
\hline
\multirow{2}{*}{\makecell{intra ratio\\($r_a$, $r_v$, $r_t$)}} 
& \multirow{2}{*}{inter} 
& \multicolumn{2}{c}{TMDC} 
& \multicolumn{2}{|c}{T2DR} 
& \multicolumn{2}{|c}{GIML} \\
\cline{3-8}
& & Acc & F1 & Acc & F1 & Acc & F1 \\
\hline
% avt
(0.0, 0.0, 0.0) & \multirow{3}{*}{AVT} & 79.07 & 79.02 & 71.34 & 71.34 & \textbf{80.82} & \textbf{80.76} \\
(0.1, 0.3, 0.5) & & 65.36 & 65.46 & 62.54 & 62.62 & \textbf{67.55} & \textbf{67.55} \\
(0.5, 0.3, 0.1) & & 76.97 & 77.01 & 70.82 & 70.89 & \textbf{77.67} & \textbf{77.60} \\
\hline
% av
(0.0, 0.0, --) & \multirow{3}{*}{AV} & 53.10 & 53.22 & 51.88 & 51.99 & \textbf{54.29} & \textbf{54.05} \\
(0.1, 0.5, --) & & 51.04 & 51.15 & \textbf{54.55} & \textbf{54.63} & 52.83 & 50.61 \\
(0.5, 0.1, --) & & 49.01 & 46.65 & 47.93 & 47.76 & \textbf{53.56} & \textbf{53.67} \\
\hline
% at
(0.0, --, 0.0) & \multirow{3}{*}{AT} & 79.24 & 79.23 & 76.50 & 76.53 & \textbf{81.11} & \textbf{81.08} \\
(0.1, --, 0.5) & & 66.94 & 66.95 & 66.01 & 66.08 & \textbf{69.88} & \textbf{69.95} \\
(0.5, --, 0.1) & & 76.09 & 76.16 & 71.60 & 71.61 & \textbf{78.78} & \textbf{78.63} \\
\hline
% vt
(--, 0.0, 0.0) & \multirow{3}{*}{VT} & 79.30 & 79.25 & 75.92 & 75.82 & \textbf{80.85} & \textbf{80.80} \\
(--, 0.1, 0.5) & & 66.01 & 66.08 & 62.45 & 62.49 & \textbf{68.54} & \textbf{68.61} \\
(--, 0.5, 0.1) & & 77.49 & 77.48 & 74.61 & 74.63 & \textbf{78.57} & \textbf{78.55} \\
\hline
% a
(0.1, --, --) & \multirow{2}{*}{A} & 52.02 & 49.81 & 54.46 & 54.51 & \textbf{55.45} & \textbf{55.49} \\
(0.5, --, --) & & 50.27 & 42.80 & 50.54 & 49.70 & \textbf{54.11} & \textbf{54.19} \\
\hline
% v
(--, 0.1, --) & \multirow{2}{*}{V} & 51.28 & 51.37 & 51.43 & 51.36 & \textbf{52.65} & \textbf{52.56} \\
(--, 0.5, --) & & 49.80 & 49.28 & 51.25 & \textbf{51.27} & \textbf{51.92} & 49.56 \\
\hline
% t
(--, --, 0.1) & \multirow{2}{*}{T} & 77.11 & 77.15 & 76.79 & 76.83 & \textbf{78.45} & \textbf{78.41} \\
(--, --, 0.5) & & 66.28 & 66.33 & 66.64 & 66.71 & \textbf{68.48} & \textbf{68.46} \\
\hline
\end{tabular}
% \vspace{-4mm}
\end{table}

Table \ref{tab:experiments_av_mask} reports results on CREMA-D and Kinetics-Sounds under varying intra-modality and inter-modality missing. Across the full degradation spectrum, from fully available modalities to completely missing ones, GIML consistently outperforms TMDC\cite{zhuang2025tmdc} and T2DR~\cite{lin2025t2dr}. On CREMA-D with fully available modalities $(0.0,0.0)$, GIML achieves 73.94\% accuracy compared to 66.99\% for TMDC and 67.93\% for T2DR. Even under severe intra-modality missing $(0.3,0.7)$, it maintains a clear advantage. A similar trend is observed on Kinetics-Sounds, where GIML consistently achieves the best performance across all settings.

Under imbalanced degradation ratios, such as $(0.1,0.5)$ or $(0.5,0.1)$, GIML consistently performs best. Existing methods do not explicitly account for modality reliability during fusion, allowing low-quality modalities to interfere with decisions. In contrast, GIML estimates modality reliability at the sample level and adjusts fusion weights accordingly, suppressing unreliable modalities while emphasizing informative ones.

To evaluate generalization across different modality combinations and modality numbers, we conduct experiments on MVSA-Single (visual–text, Table \ref{tab:experiments_vt_mask}), NVGesture (RGB–Depth, Table \ref{tab:experiments_rd_mask}), and CMU-MOSI (audio–visual–text, Table \ref{tab:experiments_avt_mask}). These datasets cover both bimodal settings and a three-modality scenario. GIML consistently achieves superior or at least competitive performance across all settings. These results demonstrate that GIML generalizes well across diverse modality combinations and remains effective when scaling from two to multiple modalities, highlighting its modality-agnostic and scalable design for incomplete multimodal learning.

\subsection{Robustness to Unseen Noise Intensity}

\begin{table}[t]
\caption{\footnotesize Performance on CREMA-D and KS under unseen continuous intra-modality and inter-modality degradation levels. Best results are highlighted in bold.}
% \vspace{-2mm}
\label{tab:experiments_av_robust}
\centering
% \footnotesize
\scriptsize
\setlength{\tabcolsep}{1pt}
\renewcommand{\arraystretch}{1.2}

\begin{tabular}{
>{\centering\arraybackslash}p{0.8cm}
|>{\centering\arraybackslash}p{1.2cm}
|>{\centering\arraybackslash}p{0.6cm}
|>{\centering\arraybackslash}p{0.9cm}
>{\centering\arraybackslash}p{0.9cm}
|>{\centering\arraybackslash}p{0.9cm}
>{\centering\arraybackslash}p{0.9cm}
|>{\centering\arraybackslash}p{0.9cm}
>{\centering\arraybackslash}p{0.9cm}
}

\hline
\multirow{2}{*}{Dataset}
& \multirow{2}{*}{\makecell{intra ratio\\($r_a$, $r_v$)}}
& \multirow{2}{*}{inter}
& \multicolumn{2}{c|}{TMDC}
& \multicolumn{2}{c|}{T2DR}
& \multicolumn{2}{c}{GIML} \\
\cline{4-9}
& & & Acc & F1 & Acc & F1 & Acc & F1 \\
\hline

% ================= CREMAD-D =================
\multirow{10}{*}{\rotatebox{90}{CREMAD-D}}

& (0.0,0.0) & \multirow{6}{*}{AV}
& 65.81 & 65.79 & 64.27 & 64.42 & \textbf{72.47} & \textbf{72.43} \\
& (0.5,0.5) &
& 61.40 & 61.25 & 61.16 & 61.43 & \textbf{66.17} & \textbf{66.45} \\
& (0.1,0.5) &
& 62.07 & 61.98 & 59.65 & 59.80 & \textbf{67.80} & \textbf{67.70} \\
& (0.5,0.1) &
& 65.54 & 65.34 & 64.14 & 64.39 & \textbf{71.36} & \textbf{71.40} \\
& (0.3,0.7) &
& 58.60 & 58.68 & 58.52 & 58.67 & \textbf{65.68} & \textbf{65.65} \\
& (0.7,0.3) &
& 63.74 & 63.68 & 62.15 & 62.55 & \textbf{66.66} & \textbf{66.66} \\
\cline{2-9}

& (0.0,-) & \multirow{2}{*}{A}
& 52.42 & 52.41 & 51.80 & 52.25 & \textbf{54.89} & \textbf{54.59} \\
& (0.5,-) &
& 49.78 & 49.80 & \textbf{51.26} & \textbf{51.53} & 51.93 & 51.68 \\
\cline{2-9}

& (--,0.0) & \multirow{2}{*}{V}
& 44.11 & 41.88 & 42.93 & 41.07 & \textbf{60.46} & \textbf{59.98} \\
& (--,0.5) &
& 41.29 & 40.04 & 39.89 & 39.16 & \textbf{55.65} & \textbf{55.40} \\
\hline
\hline
% ================= KS =================
\multirow{10}{*}{\rotatebox{90}{KS}}

& (0.0,0.0) & \multirow{6}{*}{AV}
& 58.11 & 58.34 & 58.32 & 58.37 & \textbf{69.94} & \textbf{69.53} \\
& (0.5,0.5) &
& 57.44 & 57.63 & 57.60 & 57.53 & \textbf{63.16} & \textbf{62.68} \\
& (0.1,0.5) &
& 56.40 & 56.86 & 56.08 & 55.89 & \textbf{65.31} & \textbf{64.86} \\
& (0.5,0.1) &
& 59.08 & 59.11 & 58.36 & 58.24 & \textbf{66.29} & \textbf{65.87} \\
& (0.3,0.7) &
& 55.70 & 56.11 & 53.31 & 53.31 & \textbf{61.80} & \textbf{61.40} \\
& (0.7,0.3) &
& 57.81 & 57.79 & 54.93 & 54.85 & \textbf{63.55} & \textbf{63.35} \\
\cline{2-9}

& (0.0,-) & \multirow{2}{*}{A}
& 38.25 & 38.38 & 39.45 & 39.48 & \textbf{45.78} & \textbf{45.33} \\
& (0.5,-) &
& 36.83 & 36.82 & 37.35 & 37.00 & \textbf{41.27} & \textbf{40.84} \\
\cline{2-9}

& (--,0.0) & \multirow{2}{*}{V}
& 45.99 & 44.32 & 42.96 & 41.01 & \textbf{54.39} & \textbf{53.79} \\
& (--,0.5) &
& 42.65 & 41.29 & 39.77 & 38.46 & \textbf{49.64} & \textbf{48.93} \\
\hline

\end{tabular}
% \vspace{-1mm}
\end{table}

To evaluate robustness to unseen noise intensities, the model is trained on a subset of corruption levels and evaluated on previously unseen ones. Specifically, we train the models using mask rates of 0.2, 0.4, and 0.6, and test them on unseen intensities of 0.1, 0.3, 0.5, and 0.7. The results are reported in Table~\ref{tab:experiments_av_robust} and can be compared with Table~\ref{tab:experiments_av_mask}.

Across both CREMA-D and KS, GIML exhibits only marginal performance degradation on unseen noise levels and consistently maintains superior or competitive performance compared with TMDC and T2DR. In contrast, TMDC and T2DR show noticeable performance drops when evaluated under corruption intensities not observed during training, particularly under severe intra-modality degradation or imbalanced missing ratios. This is because TMDC and T2DR implicitly treat corruption levels as discrete conditions, which restricts their ability to generalize beyond the training noise configurations. In contrast, GIML models modality degradation as a continuous variable \(\hat{\eta}\) and explicitly learns the mapping from degradation severity to modality reliability. As a result, the model can naturally interpolate to unseen noise intensities and dynamically adjust fusion weights according to modality quality, leading to more stable performance across continuous degradation scenarios.

We further examine whether \(\hat{\eta}\) captures relative degradation severity. As reported in Table~\ref{tab:nqe_trend}, \(\hat{\eta}\) exhibits strong Spearman correlation with \(\Delta f\), the cosine distance between clean and corrupted features, on mask-trained models under Gaussian corruptions. This confirms that \(\hat{\eta}\) reflects continuous degradation levels, providing a rationale for the model's robust performance on unseen intensities. This also partially accounts for GIML's robustness to noise type specificity.

\subsection{Generalization to Unseen Noise Types}
% 跨噪声类型
\begin{table}[t]
\caption{\footnotesize Performance comparison under unseen noise types (Gaussian) for audio-visual datasets.  Best results are highlighted in bold.}
% \vspace{-1mm}
\label{tab:experiments_av_guassian}
\centering
% \footnotesize
\scriptsize
\setlength{\tabcolsep}{3pt}
\renewcommand{\arraystretch}{1.2}
\begin{tabular}{c c|ccc ccc}

\hline
\multirow{2}{*}{inter} & \multirow{2}{*}{\makecell{intra ratio \\($r_a,r_v$)}}
& \multicolumn{3}{c}{CREMAD-D} 
& \multicolumn{3}{c}{KS} \\
\cline{3-8}
& & TMDC & T2DR & GIML & TMDC & T2DR & GIML \\
\hline
\multirow{4}{*}{AV} & (0,0) & 66.99 & 67.93 & \textbf{74.18} & 60.40 & 62.73 & \textbf{69.88} \\    % clean 
& (2,2) & 37.85 & 35.75 & \textbf{54.08} & 42.62 & 36.07 & \textbf{58.30} \\   % low, low
& (2,5) & 45.83 & 14.11 & \textbf{61.82} & 35.47 & 27.21 & \textbf{58.06} \\   % low medium
& (5,2) & 26.91 & 16.40 & \textbf{36.36} & 31.50 & 26.39 & \textbf{47.30} \\   % medium low
\hline
\multirow{2}{*}{A} & (0,--) & 52.66 & 53.92 & \textbf{55.57} & 38.67 & 41.21 & \textbf{45.13} \\
& (2,--) & 21.48 & 19.76 & \textbf{26.36} & 28.05 & 21.69 & \textbf{34.89} \\   % low 
\hline
\multirow{2}{*}{V} & (--,0) & 46.80 & 48.95 & \textbf{61.82} & 46.43 & 44.90 & \textbf{53.92} \\
& (--,2) & 45.56 & 38.31 & \textbf{62.36} & 27.50 & 24.61 & \textbf{35.66} \\   % low 
\hline
\end{tabular}
% \vspace{-1mm}
\end{table}

We evaluate cross-noise generalization by training on Mask corruption and testing on Gaussian noise (Table \ref{tab:experiments_av_guassian}). The Gaussian noise is zero-mean with variances denoted as $r_a,r_v$. 

All methods degrade under this distribution shift, with GIML showing the smallest drop, TMDC moderate, and T2DR the largest. The difference stems from representation handling: T2DR and TMDC extract features directly from corrupted inputs, making semantics sensitive to noise type changes. TMDC partially mitigates this via VIB compression. GIML explicitly disentangles semantic features from degradation and uses modality reliability to guide fusion, preventing noise-induced shifts from affecting semantic representations and ensuring robust performance under unseen noise types.

We further evaluate GIML on realistic-style corruptions including fog, rain, snow, and mixed Mask+Gaussian (M+G). Since existing real-world blur/defocus datasets lack aligned audio-visual emotion labels, we synthesize these corruptions on CREMA-D. Table~\ref{tab:realistic_noise} shows that GIML outperforms TMDC in most settings.

\begin{table}[ht]
\centering
\setlength{\tabcolsep}{2.8pt}
% \vspace{-4mm}
\caption{\footnotesize Accuracy (\%) under unseen corruptions on CREMA-D.}
% \vspace{-2mm}
\scriptsize
\label{tab:realistic_noise}
\begin{tabular}{lcccccccccc}
\toprule
 & \multicolumn{3}{c}{Fog} 
 & \multicolumn{3}{c}{Rain} 
 & \multicolumn{3}{c}{Snow}
 & \multicolumn{1}{c}{M+G} \\
\cmidrule(lr){2-4}
\cmidrule(lr){5-7}
\cmidrule(lr){8-10}
\cmidrule(lr){11-11}
Intensity 
& 2 & 5 & 7
& 2 & 5 & 7
& 2 & 5 & 7
& (2,2) \\
\midrule
TMDC 
& 65.29 & 57.50 & 53.81
& 64.73 & 59.24 & 55.72
& 67.12 & 64.83 & \textbf{58.22}
& 26.74 \\
GIML 
& \textbf{71.82} & \textbf{62.95} & \textbf{57.76}
& \textbf{70.37} & \textbf{62.17} & \textbf{57.50}
& \textbf{73.81} & \textbf{69.38} & 56.29
& \textbf{52.95} \\
\bottomrule
\end{tabular}
\end{table}
% \vspace{-6mm}

\subsection{Ablation Study}

\textbf{$\beta_1$ Ablation.} We evaluate the effect of $\beta_1$, which balances unimodal classification losses $\mathcal{L}{cls}^{(v)}$ and fusion loss $\mathcal{L}{cls}^{f}$. A higher $\beta_1$ emphasizes unimodal supervision, enhancing disentanglement, while a lower $\beta_1$ favors the fusion objective. Table \ref{tab:experiments_av_gamma} shows that performance remains stable across a reasonable range of $\beta_1$, indicating that the disentanglement framework effectively isolates semantic features and maintains robust fusion under noisy conditions.

\begin{table}[t]
\centering
\setlength{\tabcolsep}{4pt}
\renewcommand{\arraystretch}{0.98}
\footnotesize

\caption{\footnotesize Spearman correlations on CREMA-D. A: Audio, V: Visual}
% \vspace{-2mm}
\label{tab:nqe_trend}
\setlength{\tabcolsep}{3pt}
\begin{tabular}{lc}
\toprule
Metric & Mask / Gaussian \\
\midrule
A \(\rho(\hat{\eta}, \Delta f)\) & 0.991 / 0.973 \\
V \(\rho(\hat{\eta}, \Delta f)\) & 0.991 / 0.955 \\
A \(\rho(\mathrm{var}, \Delta f)\) & 0.927 / 0.982 \\
V \(\rho(\mathrm{var}, \Delta f)\) & 1.000 / 0.891 \\
\bottomrule
\end{tabular}
\end{table}

\begin{table}[t]
\centering
\setlength{\tabcolsep}{4pt}
\renewcommand{\arraystretch}{0.98}
\footnotesize

\caption{\footnotesize GIML performance under varying $\beta_1$ on CREMA-D across different intra-modality missing ratios.}
% \vspace{-2mm}
\label{tab:experiments_av_gamma}
\scriptsize
\setlength{\tabcolsep}{3pt}
\renewcommand{\arraystretch}{1.3}
\begin{tabular}{
>{\centering\arraybackslash}p{1.2cm}
|>{\centering\arraybackslash}p{0.6cm}
|>{\centering\arraybackslash}p{0.9cm}
>{\centering\arraybackslash}p{0.9cm}
>{\centering\arraybackslash}p{0.9cm}
>{\centering\arraybackslash}p{0.9cm}
}
\hline
\multirow{2}{*}{\makecell{intra\\($r_a$, $r_v$)}}
& \multirow{2}{*}{inter}
& \multicolumn{4}{c}{$\beta_{1}$} \\
\cline{3-6}
& & 2.0 & 3.0 & 4.0 & 5.0 \\
\hline
(0.0,0.0) & \multirow{2}{*}{AV} & 72.45 &  71.16 & \textbf{73.94} & 71.72 \\
(0.5,0.5) & & 65.59 & 65.56 & \textbf{67.80} & 66.51 \\
% (0.1,0.5) &  & 67.34 & 66.53 & 68.18 & \textbf{68.74} \\
% (0.5,0.1) &  & 70.24 & 71.21  & \textbf{72.64} & 69.25 \\
\cline{1-6}
(0.0,--) & \multirow{2}{*}{A} & 52.55 & 52.42 & \textbf{55.57} & 54.70 \\
(0.5,--) & & 50.24 & 50.19 & 50.24 & \textbf{50.89}  \\
\cline{1-6}
(--,0.0) & \multirow{2}{*}{V} & 60.38 & 61.67 & 60.38 & \textbf{61.77} \\
(--,0.5) &  & 51.75 & 53.55 & \textbf{55.30} & 53.60 \\
\hline
\end{tabular}

% \vspace{-1mm}
\end{table}

\noindent \textbf{Module Ablation.} We investigate the contributions of the modality quality estimation (NQE) and semantic disentanglement (NSD) modules. 
\begin{itemize}
    \item For GIML-NQE, the learned quality weights are disabled by setting $w^{(v)}=1$. Table \ref{tab:experiments_av_NQE} reports results under varying modality missing rates. Using uniform weights slightly reduces performance when missing rates are unbalanced, indicating that NQE improves the reliability of modality weighting. 
    
    \item For GIML-NSD, semantic disentanglement is removed by feeding the raw features $z^{(v)}$ directly into the fusion module instead of the disentangled semantic features $s^{(v)}$. Table \ref{tab:experiments_av_NSD} shows results under Gaussian noise, where this causes modest reductions in generalization and discriminability, highlighting that NSD facilitates extraction of robust semantic features for multimodal fusion. We further analyze the semantic-uncertainty decoupling of NSD. For semantics, under A/V missing rates of 0.5, we compute ``Sep.=Inter./Intra.'' on corrupted features, where ``Intra.'' is the sample-to-class-center distance and ``Inter.'' is the class-center distance. A larger ``Sep.'' indicates clearer class structure, and Table~\ref{tab:nsd_geometry} shows that \(\mu\) is more class-structured than the raw feature \(z\). For uncertainty, Table~\ref{tab:nqe_trend} shows that variance correlates with \(\Delta f\).

\end{itemize}

\begin{table}[t]
\caption{\footnotesize Impact of modality quality estimation on GIML performance under varying intra-modality and inter-modality missing on two datasets.}
% \vspace{-1mm}
\footnotesize
\label{tab:experiments_av_NQE}
\centering
\setlength{\tabcolsep}{5pt}
\renewcommand{\arraystretch}{1.2}

\begin{tabular}{c|cc cc}
\hline
\multirow{2}{*}{\makecell{intra ratio \\ $(r_a,r_v)$}}
& \multicolumn{2}{c}{CREMAD-D} 
& \multicolumn{2}{c}{KS} \\
\cline{2-5}
& GIML & GIML-NQE & GIML & GIML-NQE \\
\hline

(0.0,0.0) & \textbf{73.94} & 72.66 & \textbf{69.51} & 67.45 \\
(0.1,0.5) & \textbf{68.18} & 68.15 & \textbf{65.20} & 61.96 \\
(0.5,0.1) & \textbf{72.64} & 67.69 & \textbf{66.76} & 65.29 \\
(0.3,0.7) & 62.93 & \textbf{63.68} & \textbf{61.68} & 57.73 \\
(0.7,0.3) & \textbf{67.34} & 63.39 & \textbf{63.47} & 62.40 \\

\hline
\end{tabular}
% \vspace{-1mm}
\end{table}

\begin{table}[t]
\caption{\footnotesize Effect of semantic disentanglement on GIML generalization under Gaussian noise across different intra-modality and inter-modality missing on two datasets.}
% \vspace{-1mm}
\label{tab:experiments_av_NSD}
\centering
\scriptsize
\setlength{\tabcolsep}{6pt}
\renewcommand{\arraystretch}{1.3}
\begin{tabular}{c c|cc cc}

\hline
\multirow{2}{*}{inter} & \multirow{2}{*}{\makecell{intra ratio \\ $(r_a,r_v)$}}
& \multicolumn{2}{c}{CREMAD-D} 
& \multicolumn{2}{c}{KS} \\
\cline{3-6}
& & GIML & GIML-NSD & GIML & GIML-NSD \\
\hline
\multirow{4}{*}{AV} & (0,0) & \textbf{74.18} & 73.92 & \textbf{69.88} & 68.68 \\    % clean 
& (2,2) & \textbf{54.08} & 48.95 & \textbf{58.30} & 52.12 \\   % low, low
& (2,5) & \textbf{61.82} & 45.56 & \textbf{58.06} & 39.39 \\   % low medium
& (5,2) & \textbf{36.36} & 23.66 & \textbf{47.30} & 44.36 \\   % medium low
\hline
\multirow{2}{*}{A} & (0,--) & \textbf{55.57} & 54.57 & \textbf{45.13} & 45.11 \\
& (2,--) & \textbf{26.36} & 18.82 & \textbf{45.23} & 35.76 \\   % low 
\hline
\multirow{2}{*}{V} & (--,0) & 61.82 & \textbf{62.37} & 53.92 & \textbf{55.85} \\
& (--,2) & \textbf{62.36} & 45.56 & 35.66 & \textbf{36.99} \\   % low 
\hline
\end{tabular}
% \vspace{-2mm}
\end{table}

\begin{table}[t]
\centering
% \vspace{-3mm}
\setlength{\tabcolsep}{4pt}

\captionof{table}{\footnotesize Semantic separability under A/V missing rates of 0.5.}
% \vspace{-2mm}
\footnotesize
\label{tab:nsd_geometry}
\setlength{\tabcolsep}{6pt}
\begin{tabular}{lcc}
\toprule
Modality & \(z\) Sep. & \(\mu\) Sep. \\
\midrule
Audio  & 0.70 & \textbf{1.51} \\
Visual & 0.66 & \textbf{1.88} \\
\bottomrule
\end{tabular}
\end{table}

\begin{table}[t]
\centering
\setlength{\tabcolsep}{4pt}
\captionof{table}{\footnotesize Component ablation under Mask corruption. ``Acc.'' is at A/V missing rate 0.5. $\rho$ means Spearman correlation. Each reports A / V.}
% \vspace{-2mm}
\footnotesize
\label{tab:loss_ablation}
\begin{tabular}{lccc}  % 删掉 resizebox
\toprule
Variant & Acc. & \(\rho(\mathrm{var}, \Delta f)\) & \(\rho(\hat{\eta}, \Delta f)\) \\
\midrule
Full & \textbf{67.41} & 0.927 / 1.000 & 0.991 / 0.991 \\
w/o \(L_{reg}\) & 65.83 & 0.336 / 0.273 & 1.000 / 1.000 \\
w/o \(L_{mse}\) & 64.01 & 1.000 / 1.000 & 0.018 / 0.527 \\
\bottomrule
\end{tabular}
% \vspace{-4mm}
\end{table}

\noindent \textbf{Roles of \(L_{reg}\) and \(L_{mse}\).} We conduct component ablations under Mask corruption. As shown in Table~\ref{tab:loss_ablation}, the full model achieves the best ``Acc.'', showing that the two losses are related but not redundant. \(\Delta f\) is the cosine distance between clean and corrupted features. Removing \(L_{mse}\) mainly weakens the \(\hat{\eta}\)-\(\Delta f\) correlation, while removing \(L_{reg}\) mainly weakens the var-\(\Delta f\) correlation. Thus, \(L_{mse}\) calibrates NQE, whereas \(L_{reg}\) regularizes the uncertainty.

\section{Conclusion}

This paper introduces General Incomplete Multimodal Learning (GIML), a unified framework for multimodal learning under diverse forms of modality incompleteness. While existing methods mainly focus on inter-modality missing, and a few studies consider intra-modality corruption, they typically treat the two problems separately. In contrast, GIML models modality conditions from mild noise to complete absence as a continuous degradation spectrum, enabling a unified treatment of both intra-modality degradation and inter-modality missing. To support this formulation, we develop a Noise-aware Quality Estimator (NQE) for accurate degradation estimation and a Noise–Semantic Decoupled (NSD) module to improve robustness to diverse and unseen noise patterns. Extensive experiments demonstrate that the proposed framework achieves robust and generalizable multimodal learning across diverse degradation scenarios.

{
    \small
    \bibliographystyle{ieeenat_fullname}
    \bibliography{main}

@String(CVPR= {IEEE Conf. Comput. Vis. Pattern Recog.})

@String(AAAI = {AAAI})

@inproceedings{sun2024generative,
  title={Generative multimodal models are in-context learners},
  author={Sun, Quan and Cui, Yufeng and Zhang, Xiaosong and Zhang, Fan and Yu, Qiying and Wang, Yueze and Rao, Yongming and Liu, Jingjing and Huang, Tiejun and Wang, Xinlong},
  booktitle={Proceedings of the IEEE/CVF Conference on Computer Vision and Pattern Recognition},
  pages={14398--14409},
  year={2024}
}

@article{liang2021multibench,
  title={Multibench: Multiscale benchmarks for multimodal representation learning},
  author={Liang, Paul Pu and Lyu, Yiwei and Fan, Xiang and Wu, Zetian and Cheng, Yun and Wu, Jason and Chen, Leslie and Wu, Peter and Lee, Michelle A and Zhu, Yuke and others},
  journal={Advances in neural information processing systems},
  volume={2021},
  number={DB1},
  pages={1},
  year={2021}
}

@article{li2021dynamic,
  title={Dynamic-hierarchical attention distillation with synergetic instance selection for land cover classification using missing heterogeneity images},
  author={Li, Xiao and Lei, Lin and Sun, Yuli and Kuang, Gangyao},
  journal={IEEE Transactions on Geoscience and Remote Sensing},
  volume={60},
  pages={1--16},
  year={2021},
  publisher={IEEE}
}

@inproceedings{stroud2020d3d,
  title={D3d: Distilled 3d networks for video action recognition},
  author={Stroud, Jonathan and Ross, David and Sun, Chen and Deng, Jia and Sukthankar, Rahul},
  booktitle={Proceedings of the IEEE/CVF winter conference on applications of computer vision},
  pages={625--634},
  year={2020}
}

@article{pinto2015face,
  title={Face spoofing detection through visual codebooks of spectral temporal cubes},
  author={Pinto, Allan and Pedrini, Helio and Schwartz, William Robson and Rocha, Anderson},
  journal={IEEE Transactions on image processing},
  volume={24},
  number={12},
  pages={4726--4740},
  year={2015},
  publisher={IEEE}
}

@inproceedings{zhang2025incomplete,
  title={Incomplete Multi-view Clustering via Diffusion Contrastive Generation},
  author={Zhang, Yuanyang and Lin, Yijie and Yan, Weiqing and Yao, Li and Wan, Xinhang and Li, Guangyuan and Zhang, Chao and Ke, Guanzhou and Xu, Jie},
  booktitle={Proceedings of the AAAI Conference on Artificial Intelligence},
  volume={39},
  number={21},
  pages={22650--22658},
  year={2025}
}

@inproceedings{dai2025unbiased,
  title={Unbiased missing-modality multimodal learning},
  author={Dai, Ruiting and Li, Chenxi and Yan, Yandong and Mo, Lisi and Qin, Ke and He, Tao},
  booktitle={Proceedings of the IEEE/CVF International Conference on Computer Vision},
  pages={24507--24517},
  year={2025}
}

@article{zhang2024towards,
  title={Towards robust multimodal sentiment analysis with incomplete data},
  author={Zhang, Haoyu and Wang, Wenbin and Yu, Tianshu},
  journal={Advances in Neural Information Processing Systems},
  volume={37},
  pages={55943--55974},
  year={2024}
}

@inproceedings{zhu2025proxy,
  title={Proxy-driven robust multimodal sentiment analysis with incomplete data},
  author={Zhu, Aoqiang and Hu, Min and Wang, Xiaohua and Yang, Jiaoyun and Tang, Yiming and An, Ning},
  booktitle={Proceedings of the 63rd Annual Meeting of the Association for Computational Linguistics (Volume 1: Long Papers)},
  pages={22123--22138},
  year={2025}
}

@inproceedings{li2025simmlm,
  title={Simmlm: A simple framework for multi-modal learning with missing modality},
  author={Li, Sijie and Chen, Chen and Han, Jungong},
  booktitle={Proceedings of the IEEE/CVF International Conference on Computer Vision},
  pages={24068--24077},
  year={2025}
}

@inproceedings{sun2024redcore,
  title={RedCore: Relative advantage aware cross-modal representation learning for missing modalities with imbalanced missing rates},
  author={Sun, Jun and Zhang, Xinxin and Han, Shoukang and Ruan, Yu-Ping and Li, Taihao},
  booktitle={Proceedings of the AAAI Conference on Artificial Intelligence},
  volume={38},
  number={13},
  pages={15173--15182},
  year={2024}
}

@inproceedings{wei2023mmanet,
  title={Mmanet: Margin-aware distillation and modality-aware regularization for incomplete multimodal learning},
  author={Wei, Shicai and Luo, Chunbo and Luo, Yang},
  booktitle={Proceedings of the IEEE/CVF Conference on Computer Vision and Pattern Recognition},
  pages={20039--20049},
  year={2023}
}

@inproceedings{wang2023learnable,
  title={Learnable cross-modal knowledge distillation for multi-modal learning with missing modality},
  author={Wang, Hu and Ma, Congbo and Zhang, Jianpeng and Zhang, Yuan and Avery, Jodie and Hull, Louise and Carneiro, Gustavo},
  booktitle={International Conference on Medical Image Computing and Computer-Assisted Intervention},
  pages={216--226},
  year={2023},
  organization={Springer}
}

@inproceedings{sikdar2025ogp,
  title={OGP-Net: Optical Guidance Meets Pixel-Level Contrastive Distillation for Robust Multi-Modal and Missing Modality Segmentation},
  author={Sikdar, Aniruddh and Teotia, Jayant and Sundaram, Suresh},
  booktitle={Proceedings of the AAAI Conference on Artificial Intelligence},
  volume={39},
  number={7},
  pages={6922--6930},
  year={2025}
}

@inproceedings{lin2025t2dr,
  title={T2DR: A Two-Tier Deficiency-Resistant Framework for Incomplete Multimodal Learning},
  author={Lin, Han and Tang, Xiu and Li, Huan and Cao, Wenxue and Wu, Sai and Yao, Chang and Shou, Lidan and Chen, Gang},
  booktitle={Findings of the Association for Computational Linguistics: ACL 2025},
  pages={8602--8616},
  year={2025}
}

@article{zhuang2025tmdc,
  title={TMDC: A Two-Stage Modality Denoising and Complementation Framework for Multimodal Sentiment Analysis with Missing and Noisy Modalities},
  author={Zhuang, Yan and Liu, Minhao and Zhang, Yanru and Deng, Jiawen and Ren, Fuji},
  journal={arXiv preprint arXiv:2511.10325},
  year={2025}
}

@inproceedings{wei2024robust,
  title={Robust multimodal learning via representation decoupling},
  author={Wei, Shicai and Luo, Yang and Wang, Yuji and Luo, Chunbo},
  booktitle={European Conference on Computer Vision},
  pages={38--54},
  year={2024},
  organization={Springer}
}

@inproceedings{shi2019probabilistic,
  title={Probabilistic face embeddings},
  author={Shi, Yichun and Jain, Anil K},
  booktitle={Proceedings of the IEEE/CVF international conference on computer vision},
  pages={6902--6911},
  year={2019}
}

@article{wu2025beyond,
  title={Beyond Simple Fusion: Adaptive Gated Fusion for Robust Multimodal Sentiment Analysis},
  author={Wu, Han and Sun, Yanming and Yang, Yunhe and Wong, Derek F},
  journal={arXiv preprint arXiv:2510.01677},
  year={2025}
}

@article{li2024crossfuse,
  title={CrossFuse: A novel cross attention mechanism based infrared and visible image fusion approach},
  author={Li, Hui and Wu, Xiao-Jun},
  journal={Information Fusion},
  volume={103},
  pages={102147},
  year={2024},
  publisher={Elsevier}
}

@article{lv2025rethinking,
  title={Rethinking Cross-Modal Interaction in Multimodal Diffusion Transformers},
  author={Lv, Zhengyao and Pan, Tianlin and Si, Chenyang and Chen, Zhaoxi and Zuo, Wangmeng and Liu, Ziwei and Wong, Kwan-Yee K},
  journal={arXiv preprint arXiv:2506.07986},
  year={2025}
}

@inproceedings{cao2023multi,
  title={Multi-modal gated mixture of local-to-global experts for dynamic image fusion},
  author={Cao, Bing and Sun, Yiming and Zhu, Pengfei and Hu, Qinghua},
  booktitle={Proceedings of the IEEE/CVF international conference on computer vision},
  pages={23555--23564},
  year={2023}
}

@article{han2024fusemoe,
  title={Fusemoe: Mixture-of-experts transformers for fleximodal fusion},
  author={Han, Xing and Nguyen, Huy and Harris, Carl and Ho, Nhat and Saria, Suchi},
  journal={Advances in Neural Information Processing Systems},
  volume={37},
  pages={67850--67900},
  year={2024}
}

@inproceedings{huai2025cl,
  title={CL-MoE: Enhancing Multimodal Large Language Model with Dual Momentum Mixture-of-Experts for Continual Visual Question Answering},
  author={Huai, Tianyu and Zhou, Jie and Wu, Xingjiao and Chen, Qin and Bai, Qingchun and Zhou, Ze and He, Liang},
  booktitle={Proceedings of the Computer Vision and Pattern Recognition Conference},
  pages={19608--19617},
  year={2025}
}

@article{jing2025evomoe,
  title={EvoMoE: Expert Evolution in Mixture of Experts for Multimodal Large Language Models},
  author={Jing, Linglin and Gao, Yuting and Wang, Zhigang and Lan, Wang and Tang, Yiwen and Wang, Wenhai and Zhang, Kaipeng and Guo, Qingpei},
  journal={arXiv preprint arXiv:2505.23830},
  year={2025}
}

@inproceedings{zeevi2025rate,
  title={Rate-In: Information-Driven Adaptive Dropout Rates for Improved Inference-Time Uncertainty Estimation},
  author={Zeevi, Tal and Shwartz-Ziv, Ravid and LeCun, Yann and Staib, Lawrence H and Onofrey, John A},
  booktitle={Proceedings of the Computer Vision and Pattern Recognition Conference},
  pages={20757--20766},
  year={2025}
}

@article{djupskaas2025unreliable,
  title={Unreliable Uncertainty Estimates with Monte Carlo Dropout},
  author={Djupsk{\aa}s, Aslak and Stasik, Alexander Johannes and Riemer-S{\o}rensen, Signe},
  journal={arXiv preprint arXiv:2512.14851},
  year={2025}
}

@article{yang2024uncertainty,
  title={Uncertainty-based offline variational bayesian reinforcement learning for robustness under diverse data corruptions},
  author={Yang, Rui and Wang, Jie and Wu, Guoping and Li, Bin},
  journal={Advances in Neural Information Processing Systems},
  volume={37},
  pages={39748--39783},
  year={2024}
}

@article{harrison2024variational,
  title={Variational Bayesian last layers},
  author={Harrison, James and Willes, John and Snoek, Jasper},
  journal={arXiv preprint arXiv:2404.11599},
  year={2024}
}

@article{venkataramanan2025probabilistic,
  title={Probabilistic Embeddings for Frozen Vision-Language Models: Uncertainty Quantification with Gaussian Process Latent Variable Models},
  author={Venkataramanan, Aishwarya and Bodesheim, Paul and Denzler, Joachim},
  journal={arXiv preprint arXiv:2505.05163},
  year={2025}
}

@article{lin2025intra,
  title={Intra-Class Probabilistic Embeddings for Uncertainty Estimation in Vision-Language Models},
  author={Lin, Zhenxiang and Haghighat, Maryam and Browne, Will and Miller, Dimity},
  journal={arXiv preprint arXiv:2511.22019},
  year={2025}
}

@inproceedings{cai2018deep,
  title={Deep adversarial learning for multi-modality missing data completion},
  author={Cai, Lei and Wang, Zhengyang and Gao, Hongyang and Shen, Dinggang and Ji, Shuiwang},
  booktitle={Proceedings of the 24th ACM SIGKDD international conference on knowledge discovery \& data mining},
  pages={1158--1166},
  year={2018}
}

@inproceedings{jue2019integrating,
  title={Integrating cross-modality hallucinated MRI with CT to aid mediastinal lung tumor segmentation},
  author={Jue, Jiang and Jason, Hu and Neelam, Tyagi and Andreas, Rimner and Sean, Berry L and Joseph, Deasy O and Harini, Veeraraghavan},
  booktitle={International conference on medical image computing and computer-assisted intervention},
  pages={221--229},
  year={2019},
  organization={Springer}
}

@article{liu2021face,
  title={Face anti-spoofing via adversarial cross-modality translation},
  author={Liu, Ajian and Tan, Zichang and Wan, Jun and Liang, Yanyan and Lei, Zhen and Guo, Guodong and Li, Stan Z},
  journal={IEEE Transactions on Information Forensics and Security},
  volume={16},
  pages={2759--2772},
  year={2021},
  publisher={IEEE}
}

@inproceedings{havaei2016hemis,
  title={Hemis: Hetero-modal image segmentation},
  author={Havaei, Mohammad and Guizard, Nicolas and Chapados, Nicolas and Bengio, Yoshua},
  booktitle={International conference on medical image computing and computer-assisted intervention},
  pages={469--477},
  year={2016},
  organization={Springer}
}

@inproceedings{zhang2022mmformer,
  title={mmformer: Multimodal medical transformer for incomplete multimodal learning of brain tumor segmentation},
  author={Zhang, Yao and He, Nanjun and Yang, Jiawei and Li, Yuexiang and Wei, Dong and Huang, Yawen and Zhang, Yang and He, Zhiqiang and Zheng, Yefeng},
  booktitle={International conference on medical image computing and computer-assisted intervention},
  pages={107--117},
  year={2022},
  organization={Springer}
}

@inproceedings{praveen2024recursive,
  title={Recursive joint cross-modal attention for multimodal fusion in dimensional emotion recognition},
  author={Praveen, R Gnana and Alam, Jahangir},
  booktitle={Proceedings of the IEEE/CVF Conference on Computer Vision and Pattern Recognition},
  pages={4803--4813},
  year={2024}
}

@inproceedings{zhuanghyper,
  title={Hyper-Modality Enhancement for Multimodal Sentiment Analysis with Missing Modalities},
  author={Zhuang, Yan and Minhao, LIU and Bai, Wei and Zhang, Yanru and Li, Wei and Deng, Jiawen and Ren, Fuji},
  booktitle={The Thirty-ninth Annual Conference on Neural Information Processing Systems}
}

@inproceedings{araujo2025cav,
  title={CAV-MAE Sync: Improving Contrastive Audio-Visual Mask Autoencoders via Fine-Grained Alignment},
  author={Araujo, Edson and Rouditchenko, Andrew and Gong, Yuan and Bhati, Saurabhchand and Thomas, Samuel and Kingsbury, Brian and Karlinsky, Leonid and Feris, Rogerio and Glass, James R and Kuehne, Hilde},
  booktitle={Proceedings of the Computer Vision and Pattern Recognition Conference},
  pages={18794--18803},
  year={2025}
}

@inproceedings{dai2024study,
  title={A study of dropout-induced modality bias on robustness to missing video frames for audio-visual speech recognition},
  author={Dai, Yusheng and Chen, Hang and Du, Jun and Wang, Ruoyu and Chen, Shihao and Wang, Haotian and Lee, Chin-Hui},
  booktitle={Proceedings of the IEEE/CVF Conference on Computer Vision and Pattern Recognition},
  pages={27445--27455},
  year={2024}
}

@article{jin2025rohydr,
  title={RoHyDR: Robust Hybrid Diffusion Recovery for Incomplete Multimodal Emotion Recognition},
  author={Jin, Yuehan and Liu, Xiaoqing and Yang, Yiyuan and Yu, Zhiwen and Zhang, Tong and Yang, Kaixiang},
  journal={arXiv preprint arXiv:2505.17501},
  year={2025}
}

@inproceedings{xu2024leveraging,
  title={Leveraging knowledge of modality experts for incomplete multimodal learning},
  author={Xu, Wenxin and Jiang, Hexin and Liang, Xuefeng},
  booktitle={Proceedings of the 32nd ACM International Conference on Multimedia},
  pages={438--446},
  year={2024}
}

@inproceedings{wang2020makes,
  title={What makes training multi-modal classification networks hard?},
  author={Wang, Weiyao and Tran, Du and Feiszli, Matt},
  booktitle={Proceedings of the IEEE/CVF conference on computer vision and pattern recognition},
  pages={12695--12705},
  year={2020}
}

@inproceedings{wei2025boosting,
  title     = {Boosting Multimodal Learning via Disentangled Gradient Learning},
  author    = {Wei, Shicai and Luo, Chunbo and Luo, Yang},
  booktitle = {Proceedings of the IEEE/CVF International Conference on Computer Vision},
  pages     = {22879--22888},
  year      = {2025}
}

@inproceedings{wei2026unbiased,
  title     = {Unbiased Dynamic Multimodal Fusion},
  author    = {Wei, Shicai and Zhang, Kaijie and Chen, Luyi and He, Tao and Duan, Guiduo},
  booktitle = {Proceedings of the IEEE/CVF Conference on Computer Vision and Pattern Recognition},
  pages     = {6239--6249},
  year      = {2026}
}

@inproceedings{peng2022balanced,
  title={Balanced multimodal learning via on-the-fly gradient modulation},
  author={Peng, Xiaokang and Wei, Yake and Deng, Andong and Wang, Dong and Hu, Di},
  booktitle={Proceedings of the IEEE/CVF conference on computer vision and pattern recognition},
  pages={8238--8247},
  year={2022}
}

@article{chaudhuri2025closer,
  title={A Closer Look at Multimodal Representation Collapse},
  author={Chaudhuri, Abhra and Dutta, Anjan and Bui, Tu and Georgescu, Serban},
  journal={arXiv preprint arXiv:2505.22483},
  year={2025}
}

@article{liu2020energy,
  title={Energy-based out-of-distribution detection},
  author={Liu, Weitang and Wang, Xiaoyun and Owens, John and Li, Yixuan},
  journal={Advances in neural information processing systems},
  volume={33},
  pages={21464--21475},
  year={2020}
}

@inproceedings{zadeh2018multi,
  title={Multi-attention recurrent network for human communication comprehension},
  author={Zadeh, Amir and Liang, Paul Pu and Poria, Soujanya and Vij, Prateek and Cambria, Erik and Morency, Louis-Philippe},
  booktitle={Proceedings of the AAAI conference on artificial intelligence},
  volume={32},
  number={1},
  year={2018}
}

@inproceedings{arandjelovic2017look,
  title={Look, listen and learn},
  author={Arandjelovic, Relja and Zisserman, Andrew},
  booktitle={Proceedings of the IEEE international conference on computer vision},
  pages={609--617},
  year={2017}
}

@article{cao2014crema,
  title={Crema-d: Crowd-sourced emotional multimodal actors dataset},
  author={Cao, Houwei and Cooper, David G and Keutmann, Michael K and Gur, Ruben C and Nenkova, Ani and Verma, Ragini},
  journal={IEEE transactions on affective computing},
  volume={5},
  number={4},
  pages={377--390},
  year={2014},
  publisher={IEEE}
}

@inproceedings{gupta2016online,
  title={Online detection and classification of dynamic hand gestures with recurrent 3d convolutional neural networks},
  author={Gupta, PMXYS and Kautz, KKSTJ and others},
  booktitle={CVPR},
  volume={1},
  number={2},
  pages={3},
  year={2016}
}

@inproceedings{niu2016sentiment,
  title={Sentiment analysis on multi-view social data},
  author={Niu, Teng and Zhu, Shiai and Pang, Lei and El Saddik, Abdulmotaleb},
  booktitle={International conference on multimedia modeling},
  pages={15--27},
  year={2016},
  organization={Springer}
}

@inproceedings{zhang2023provable,
  title={Provable dynamic fusion for low-quality multimodal data},
  author={Zhang, Qingyang and Wu, Haitao and Zhang, Changqing and Hu, Qinghua and Fu, Huazhu and Zhou, Joey Tianyi and Peng, Xi},
  booktitle={International conference on machine learning},
  pages={41753--41769},
  year={2023},
  organization={PMLR}
}

@inproceedings{he2016deep,
  title={Deep residual learning for image recognition},
  author={He, Kaiming and Zhang, Xiangyu and Ren, Shaoqing and Sun, Jian},
  booktitle={Proceedings of the IEEE conference on computer vision and pattern recognition},
  pages={770--778},
  year={2016}
}

@inproceedings{devlin2019bert,
  title={Bert: Pre-training of deep bidirectional transformers for language understanding},
  author={Devlin, Jacob and Chang, Ming-Wei and Lee, Kenton and Toutanova, Kristina},
  booktitle={Proceedings of the 2019 conference of the North American chapter of the association for computational linguistics: human language technologies, volume 1 (long and short papers)},
  pages={4171--4186},
  year={2019}
}

@inproceedings{wei2025improving,
  title={Improving multimodal learning via imbalanced learning},
  author={Wei, Shicai and Luo, Chunbo and Luo, Yang},
  booktitle={Proceedings of the IEEE/CVF International Conference on Computer Vision},
  pages={2250--2259},
  year={2025}
}

@article{wei2024gradient,
  title     = {Gradient Decoupled Learning With Unimodal Regularization for Multimodal Remote Sensing Classification},
  author    = {Wei, Shicai and Luo, Chunbo and Ma, Xiaoguang and Luo, Yang},
  journal   = {IEEE Transactions on Geoscience and Remote Sensing},
  volume    = {62},
  pages     = {1--12},
  year      = {2024},
  publisher = {IEEE},
  doi       = {10.1109/TGRS.2024.3478393}
}

@article{wei2023mshnet,
  title     = {{MSH-Net}: Modality-Shared Hallucination With Joint Adaptation Distillation for Remote Sensing Image Classification Using Missing Modalities},
  author    = {Wei, Shicai and Luo, Yang and Ma, Xiaoguang and Ren, Peng and Luo, Chunbo},
  journal   = {IEEE Transactions on Geoscience and Remote Sensing},
  volume    = {61},
  pages     = {1--15},
  year      = {2023},
  publisher = {IEEE},
  doi       = {10.1109/TGRS.2023.3265650},
  articleno = {4402615}
}
}


\end{document}